\documentclass[10pt,conference]{IEEEtran}
\IEEEoverridecommandlockouts

\usepackage{soul}
\usepackage[dvipsnames]{xcolor}

\usepackage{lettrine}

\usepackage{cite}
\usepackage{amsmath,amssymb,amsfonts}
\usepackage{algorithmic}
\usepackage{graphicx}
\usepackage{textcomp}
\usepackage{xcolor}
\usepackage{colortbl}
\usepackage{booktabs}
\usepackage{verbatim}
\usepackage{multirow}
\usepackage{caption}   
\usepackage{subcaption} 
\usepackage[T1]{fontenc}

\usepackage[colorlinks = true,
linkcolor = blue,
urlcolor  = black,
citecolor = blue,
anchorcolor = blue]{hyperref}

\usepackage{tikz}
\usetikzlibrary{shapes,arrows,positioning,calc,matrix}

\usepackage{cite}

\def\BibTeX{{\rm B\kern-.05em{\sc i\kern-.025em b}\kern-.08em
    T\kern-.1667em\lower.7ex\hbox{E}\kern-.125emX}}
    
\begin{document}


\title{Enhancing Network Security Management in Water Systems using FM-based Attack Attribution}

\author{
    \IEEEauthorblockN{
        Aleksandar Avdalović\IEEEauthorrefmark{1},
        Joseph Khoury\IEEEauthorrefmark{1},
        Ahmad Taha\IEEEauthorrefmark{2},
        Elias Bou-Harb\IEEEauthorrefmark{1}
    }
    \IEEEauthorblockA{\IEEEauthorrefmark{1}Division of Computer Science and Engineering, Louisiana State University, USA}
    \IEEEauthorblockA{\IEEEauthorrefmark{2}Civil and Environmental Engineering, Vanderbilt University, USA}
}


\maketitle

\begin{abstract}
Water systems are vital components of modern infrastructure, yet they are increasingly susceptible to sophisticated cyber attacks with potentially dire consequences on public health and safety. While state-of-the-art machine learning techniques effectively detect anomalies, contemporary model-agnostic attack attribution methods using \texttt{LIME}, \texttt{SHAP}, and \texttt{LEMNA} are deemed impractical for large-scale, interdependent water systems. This is due to the intricate interconnectivity and dynamic interactions that define these complex environments. Such methods primarily emphasize individual feature importance while falling short of addressing the crucial sensor-actuator interactions in water systems, which limits their effectiveness in identifying root cause attacks. To this end, we propose a novel model-agnostic Factorization Machines (FM)-based approach that capitalizes on water system sensor-actuator interactions to provide granular explanations and attributions for cyber attacks. For instance, an anomaly in an actuator pump activity can be attributed to a top root cause attack candidates, a list of water pressure sensors, which is derived from the underlying linear and quadratic effects captured by our approach. We validate our method using two real-world water system specific datasets, \texttt{SWaT} and \texttt{WADI}, demonstrating its superior performance over traditional attribution methods. In multi-feature cyber attack scenarios involving intricate sensor-actuator interactions, our FM-based attack attribution method effectively ranks attack root causes, achieving approximately 20\% average improvement over \texttt{SHAP} and \texttt{LEMNA}. Additionally, our approach maintains strong performance in single-feature attack scenarios, demonstrating versatility across different types of cyber attacks. Notably, our approach maintains a low computational overhead equating to an O(n) time complexity, making it suitable for real-time applications in critical water system infrastructure. Our work underscores the importance of modeling feature interactions in water systems, offering a robust tool for operators to diagnose and mitigate root cause attacks more effectively.
\end{abstract}


\section{Introduction}

\lettrine{W}{ATER} systems at the physical layer comprise critical components such as flow and pressure sensors, and actuators, which are monitored and controlled by cyber layer systems to ensure a safe and reliable water supply for both communities and industries. In the United States, these critical networks consist of over 152,000 publicly owned facilities that extend across more than than 2.2 million miles of transmission and distribution lines, with an estimated maintenance and upgrade cost of \$625 billion \cite{umich2023water}.

The essential nature of these systems makes them prime targets for nation-state actors aiming to disrupt operations and create chaos.\cite{smartcity} Recent state-sponsored attacks on water treatment plants have forced facilities to switch to manual operations to prevent disasters and control chemical levels in water supplies.\cite{kans, goodin2021florida, tucker2020israel}. Additionally, breaches targeting specific components have demonstrated the potential for widespread disruption and public safety risks \cite{kim2023iran}, \cite{eliascps}. These malicious activities highlight the urgent need for water systems-centric anomaly attribution methods that not only detect anomalies but also accurately identify their sources. Effective attribution is essential for isolating compromised components, mitigating threats, and implementing targeted defenses \cite{Fung2024}. Currently, anomaly detection frameworks often lack integrated attribution and explainability capabilities, limiting the ability to swiftly identify and respond to complex cyber-physical attacks. Addressing this gap is crucial for enhancing the security and resilience of water infrastructure against sophisticated threats \cite{moraitis2022cyberphysical, taormina2017characterizing}.

Indeed, water systems, such as water distribution networks and urban sewage systems, operate on process-level data from sensors that monitor physical processes and actuators that regulate them \cite{eliascps}. 
A common threat model for these systems involves manipulating process-level data \cite{giraldo2018survey, ha2022explainable}, classified under MITRE ATT\&CK Technique T1565\cite{mitreT1565}, which can lead to hazardous outcomes such as tank overflows, intentional water contamination, or large-scale leaks \cite{ahmed2017wadi}.
To counter these threats, various real-time manipulation detection methods have been developed, with Machine Learning (ML) increasingly adopted as a preferred approach  \cite{khoury2020hybrid, alsabaan2023,yang2022,ahmed2023,Joseph}. A common approach for anomaly detection in current systems involves using ML models that treat process-level values, such as sensor and actuator data, as features. These models are trained to predict future states of the system, performing a per-feature (or per-asset) prediction for each sensor and actuator based on an input time window of prior states\footnote{
Throughout this paper, we will use the terms \textit{\textbf{feature}} and \textit{\textbf{asset}} interchangeably. Both terms refer to \textbf{\textit{sensors and actuators}} within water systems that are central to our anomaly detection and attribution methods.}. During operation, the predicted state is compared with the observed state; if the reconstruction error exceeds a predetermined threshold, an anomaly is flagged. This ML-based framework has been shown to effectively detect anomalies across various types of water systems \cite{ahmed2017wadi, goh2016dataset}, and is applicable across different model architectures \cite{fung2022perspectives,zizzo2019intrusion}. Yet, a key challenge towards addressing anomaly detection is not only identifying that an anomaly exists but also determining its \textit{cause}. In this work, we address this challenge through two complementary objectives, namely, \textit{attribution} and \textit{explainability}. \\

\noindent \textbf{\textit{Attribution}} helps operators identify the cause of an anomaly by quantifying the contribution of each feature towards the model's prediction decision. In the context of water systems, process-level data (such as sensor and actuator readings) are directly tied to physical components of the system. Therefore, attribution methods can help pinpoint which specific sensors or actuators are responsible for the anomaly \cite{hwang2021esfd}. 
This is important for water systems because operators must have clear insights to quickly respond to both attacks and malfunctions, which can directly impact public health and safety. While model-agnostic attribution methods like \texttt{LIME} \cite{Ribeiro2016LIME}, \texttt{SHAP} \cite{lundberg2017unified}, and \texttt{LEMNA} \cite{guo2018lemna} are widely used for individual feature importance, model-specific approaches such as \texttt{Integrated Gradients} \cite{sundararajan2017axiomatic}, \texttt{SmoothGrad} \cite{smilkov2017smoothgrad}, and \texttt{Saliency Maps} \cite{simonyan2013deep} have been developed to handle more complex scenarios. However, these methods, whether model-agnostic or model-specific, often fail to capture the complex interactions between sensors and actuators that are critical to the functionality of water systems.

While attribution focuses on identifying the most impactful features, \textit{\textbf{explainability}} more broadly refers to understanding how a model arrives at its predictions. In recent years, explainability methods have gained traction across various fields, particularly in security-related ML applications. These methods provide a means to interpret the behavior of complex models, which are often referred to as ``black box'' due to the difficulty in tracing how they make decisions \cite{simonyan2013deep}. Using explainability techniques in anomaly detection models helps operators and the public trust these systems, enhancing transparency, reliability, and security management in critical infrastructure.\newline

\noindent \textbf{\textit{Challenges}.} Although attribution and explainability techniques have been widely used, applying them to water systems presents specific challenges. Traditional model-agnostic methods like \texttt{LIME} \cite{Ribeiro2016LIME}, \texttt{SHAP} \cite{lundberg2017unified}, and \texttt{LEMNA} \cite{guo2018lemna} focus on individual feature importance by providing linear approximations of feature contributions. While \texttt{SHAP} and \texttt{LEMNA} can account for some interactions, their outputs are still primarily based on independent feature contributions. In water systems, however, sensors and actuators are tightly interconnected, and understanding these interactions is crucial for identifying the root causes of anomalies \cite{Fung2024}. Neglecting these connections could cause attribution methods to overlook key factors behind complex, targeted attacks \cite{sadegh}. In terms of explainability, while deep learning models like CNNs \cite{lecun1998gradient} and LSTMs \cite{hochreiter1997long} are effective at detecting novel threats \cite{fung2022perspectives}, their complexity makes understanding their decisions a significant challenge. In time-sensitive situations, this lack of transparency reduces effectiveness, as operators need to quickly understand how decisions are made. Explainability methods that focus only on individual features, without accounting for interactions, can make it harder to diagnose system failures or instabilities. Developing explanation techniques that account for feature interactions is important for securing water systems against sophisticated threats, such as multi-asset attacks. \newline

\noindent \textit{\textbf{Scientific Contributions.}} This work addresses the challenges of attribution and explainability in water systems by using Factorization Machines (FM) \cite{Rendle2010FM}, originally developed for recommendation systems \cite{Rendle2012FMlibFM}, as a model-agnostic explainer for existing anomaly detectors. FM identifies both individual asset contributions and interactions between components, such as sensors and actuators, which can be crucial for diagnosing anomalies. By capturing these interactions, our approach offers more accurate attribution, improving the detection and mitigation of complex attacks on water infrastructure.

FM provides explainability by capturing both linear feature contributions and second-order interactions (quadratic effects), offering a more detailed and precise understanding than traditional linear methods. Additionally, FM operates efficiently in linear time \cite{Rendle2010FM}, making it ideal for real-time applications. Our approach improves interpretability and security management while retaining the underlying detection models. In brief, the paper makes the following scientific and pragmatic contributions:

\begin{itemize}

    \item We introduce a novel model-agnostic framework that leverages Factorization Machines (FM) to enhance attack attribution and explainability in water systems.
    
    

    \item We model both linear individual contributions and quadratic pairwise interaction effects in water systems using second-order FM equation, enabling more precise attack attribution while preserving computational efficiency.

    \item We validate our proposed approach as a effective plug-and-play attack attribution method that seamlessly integrates with existing deep learning anomaly detection models in water systems.

    \item We evaluated our FM-based attack attribution method using two real-world water system datasets, \texttt{SWaT} and \texttt{WADI}. The results demonstrate improved performance in accurately identifying manipulated features during complex, multi-feature attack scenarios, compared to other model-agnostic attribution techniques.

\end{itemize}

The paper is structured as follows: Section 2 presents a motivating example, illustrating how the proposed scheme captures interactions in water systems. Section 3 explains the FM-based methodology for modeling individual and interaction effects. Section 4 details the experimental setup and results. Section 5 reviews related work, and Section 6 concludes with findings, limitations, and future directions.

\begin{figure*}[!htbp]
\centering
\includegraphics[width=\linewidth]{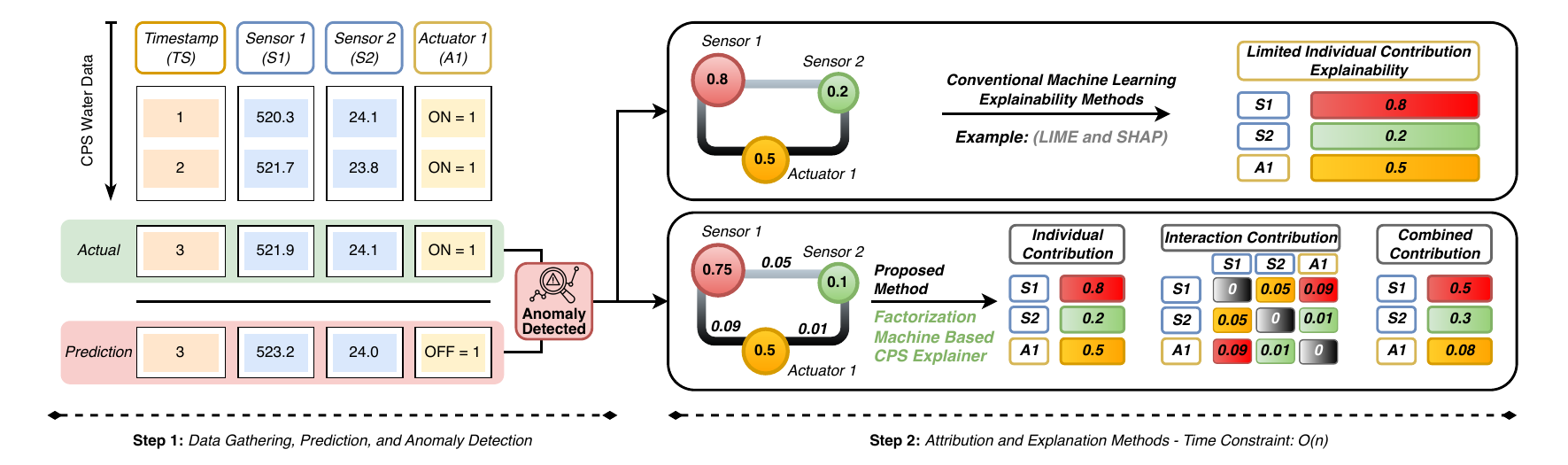}
\caption{Illustrative Scenario: (Left) Data gathering and anomaly detection in a water system, where an anomaly is detected due to a mismatch between predicted and observed values. (Upper Right) Traditional attribution methods focus on individual contributions but overlook interactions. (Lower Right) The proposed FM-based approach captures both individual contributions and interactions, providing a comprehensive understanding of the anomaly's root cause. This example is motivated by a real attack scenario further analyzed in Figure \ref{fig:combined_weights}.}
\label{motivation_example}
\end{figure*}

\section{Motivating Example}

Consider the scenario depicted in Figure ~\ref{motivation_example}, where data from a water system’s sensors and actuators is collected over multiple time steps. Machine learning models like Convolutional Neural Networks (CNNs) \cite{lecun1998gradient} and Long Short-Term Memory (LSTM) networks \cite{hochreiter1997long}, while not state-of-the-art in all domains, are widely regarded as state-of-the-art for anomaly detection in Cyber-Physical Systems (CPS) \cite{Fung2024}. In this scenario, these models analyze historical sensor and actuator data to identify patterns and predict future system states, enabling earlier and more accurate detection of anomalies. For example, an anomaly is detected at timestamp 3 due to a significant mismatch between the predicted and observed values. Traditional model-agnostic attribution methods like \texttt{LIME} \cite{Ribeiro2016LIME} and \texttt{SHAP} \cite{lundberg2017unified}, shown in the upper right of Figure ~\ref{motivation_example}, quantify the individual contributions of each feature, such as sensors or actuators. This approach overlooks the critical interplay between sensors and actuators essential for understanding anomalies \cite{Fung2024}. The proposed scheme herein leverages FM to capture both individual contributions and interactions between assets within the same time frame. As illustrated in the lower right of Figure~\ref{motivation_example}, the FM approach models the effects of interactions alongside individual asset contributions. For example, the interaction between Sensor 1 and Actuator 1 shows a notable contribution to the anomaly, a relationship that traditional methods miss. This attribution mechanism enables water plant operators to not only identify which assets are involved but also how their interactions contribute to the system malfunctions. Such insights are especially valuable in complex scenarios where component interactions, such as cascading failures \cite{LI2021204}, play a critical role \cite{antonio_elias}. By providing both individual and interaction contributions in real-time without additional computational overhead, our method equips operators with actionable insights to prevent further damage while prioritizing system recovery. Further details are provided in  
Section \ref{sec:methodology}.

\begin{figure*}[ht]
    \centering
    \includegraphics[width=\textwidth]{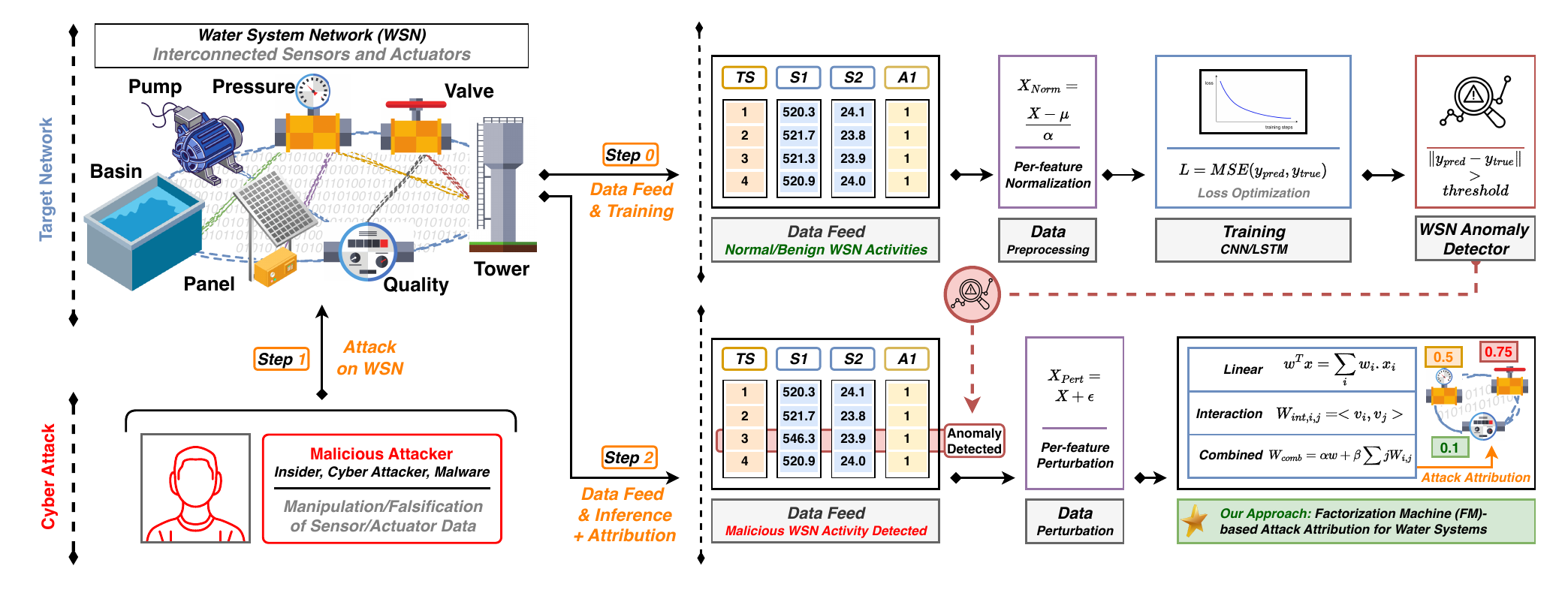}
    \caption{Overview of the Water System Anomaly Detection and Attribution Process. The diagram illustrates the steps involved in anomaly detection and attribution within a water system. The system comprises interconnected sensors and actuators (e.g., pumps, pressure valves) that feed data into a deep learning model (CNN/LSTM). This model is trained to detect anomalies based on a reconstruction error threshold. Once an anomaly is detected, the FM model performs attribution by identifying the responsible sensors and actuators, capturing both individual and interaction effects between components. The process shows data collection (Step 0), attack simulation (Step 1), and anomaly detection with attribution (Step 2). The bottom part highlights how FM combines linear and interaction terms for more accurate anomaly attribution in water systems.}
    \label{fig:methodology_overview}
\end{figure*}

\vspace{-0.2cm}

\section{Methodology}
\label{sec:methodology}

Figure ~\ref{fig:methodology_overview} provides an overview of our proposed approach. It integrates deep learning models with FM for improved anomaly detection and attribution in water systems. CNNs and LSTMs predict future system states, flagging anomalies when predictions significantly deviate from observed values. FM then identifies the contributing sensors and actuators by analyzing individual and interaction effects. After detecting anomalies, FM provides operators with a ranked list of likely responsible assets and detailed attribution coefficients for further investigation.

\subsection{Anomaly Detection with Deep Learning}

We employ deep-learning models, specifically Convolutional Neural Networks (CNNs) and Long Short-Term Memory (LSTM) networks, for anomaly detection in water systems. These models operate in an unsupervised setting, where the objective is to predict the next system state based on prior sensor and actuator data. Unsupervised learning is preferred in this context, as supervised approaches rely on explicit attack labels, which are often sparse and difficult to generalize due to the sparsity of attack data in water systems\cite{apruzzese2023machine}.
 An anomaly is flagged when the prediction error exceeds a predefined threshold. Given an input window of the previous \( h \) states, \( (x_{t-h}, \dots, x_{t-1}) \), the model predicts the next state \( \hat{x}_t \). The predicted state is compared with the observed state \( x_t \), and an anomaly is detected if the error \( \| \hat{x}_t - x_t \| \) exceeds a threshold \( \epsilon \).
CNNs capture short-term dependencies in time-series data using one-dimensional convolutional kernels. LSTMs, on the other hand, model longer-term dependencies with memory units, allowing them to track hidden states and detect subtle, delayed anomalies. Both models are trained on normal process data, learning typical system behavior. During operation, they flag anomalies when the deviation between the predicted and actual states surpasses the defined threshold.
These models were selected for their proven effectiveness in previous ICS research \cite{kravchik2022efficient,zizzo2019intrusion,fung2022perspectives,Fung2024}. CNNs are well-suited for short-term patterns, while LSTMs are better at capturing long-term relationships, enabling the detection of both immediate and evolving anomalies in the system.

\subsection{Factorization Machine Model}

FM is effective for capturing both individual feature contributions and interactions, making it suitable for tasks requiring prediction and explanation. The FM model predicts an outcome $\hat{y}$ using Equation \eqref{eq:fm_model}:

\begin{equation}
\hat{y} = b + \mathbf{w}^\top \mathbf{x} + \sum_{i=1}^{n} \sum_{j=i+1}^{n} \langle \mathbf{v}_i, \mathbf{v}_j \rangle x_i x_j
\label{eq:fm_model}
\end{equation}

\noindent where $b$ is the bias term, $\mathbf{w}$ represents vector of linear weights, and $\mathbf{v}_i$ are latent vectors capturing pairwise feature interactions. The hyperparameter $k$, the number of latent factors, is predetermined. This ensures FM operates with linear computational complexity, $O(kn)$, where $n$ is the number of features \cite{Rendle2010FM}. FM matches the computational complexity of \texttt{LIME} \cite{Ribeiro2016LIME} while additionally capturing interactions between features \cite{Rendle2010FM}.

\subsection{Training the FM Explainer}

The FM explainer is trained using perturbed samples generated at the time of anomaly detection. Let $\mathbf{X} \in \mathbb{R}^{d \times n}$ be the input data matrix, where $d$ is the number of historical samples and $n$ is the number of features. We generate perturbed samples $\mathbf{X}_{\text{pert}}$ by adding Gaussian noise as show in Equation \eqref{eq:gaussian_noise}, where $\sigma^2$ is the variance of the Gaussian noise.


\begin{equation}
\mathbf{X}_{\text{pert}} = \mathbf{X} + \epsilon, \quad \epsilon \sim \mathcal{N}(0, \sigma^2\mathbf{I})
\label{eq:gaussian_noise}
\end{equation}

\noindent The FM explainer is trained by minimizing a weighted mean squared error (MSE) loss function with L1 regularization, as provided in Equation  \eqref{eq:loss_function}.
The L1 regularization promotes sparsity in both the linear weights $\mathbf{w}$ and the interaction factors $\mathbf{V}$, helping to identify the most relevant features and interactions.

\begin{equation}
L(\mathbf{w}, \mathbf{V}) = \frac{1}{m} \sum_{i=1}^{m} w_i (\hat{y}_i - y_i)^2 + \lambda_w \|\mathbf{w}\|_1 + \lambda_V \|\mathbf{V}\|_1
\label{eq:loss_function}
\end{equation}

\noindent where $m$ is the number of perturbed samples, $w_i$ is the weight assigned to the $i$-th sample (based on its proximity to the original data point), $\hat{y}_i$ and $y_i$ are the predicted and actual values for the $i$-th sample respectively, $\lambda_w$ and $\lambda_V$ are regularization parameters, and $\|\cdot\|_1$ denotes the L1 norm. \\
The model parameters $\mathbf{w} \in \mathbb{R}^n$ and $\mathbf{V} \in \mathbb{R}^{n \times k}$ are optimized using the Adam optimizer, and $k$ is the number of latent factors. The optimization problem can be formulated as presented in Equation \eqref{eq:adam_algo}:


\begin{equation}
\mathbf{w}^*, \mathbf{V}^* = \arg\min_{\mathbf{w}, \mathbf{V}} L(\mathbf{w}, \mathbf{V})
\label{eq:adam_algo}
\end{equation}

\noindent Thus, the training process provides both linear $\mathbf{w}$ and interaction weights  $\mathbf{V}$, which can be further ranked and used to identify the most influential features and their interactions. The sparsity induced by L1 regularization focuses on the most relevant factors, improving the clarity of the attribution.

\subsection{Attribution and Explanation Mechanisms}

The FM explainer provides a precise attribution of anomalies to specific features and their interactions, offering a comprehensive explanation of the model's predictions.

\noindent \textbf{Linear Weights:} The linear weights $\mathbf{w}$ quantify the individual contribution of each feature (asset) as provided in Equation \eqref{eqn:linear_weights}, where $w_i$ represents the linear weight assigned to feature $i$, and $x_i$ is the corresponding feature value.

\begin{equation}
\mathbf{w}^\top \mathbf{x} = \sum_{i=1}^{n} w_i x_i
\label{eqn:linear_weights}
\end{equation}

\noindent This expression allows us to directly measure how much each feature, in isolation, influences the model’s output.

\noindent \textbf{Interaction Weights:} The interaction weights capture the effects of feature interactions as presented in Equation \eqref{eqn:interaction_weights}, where $\mathbf{v}_i$ and $\mathbf{v}_j$ are the latent vectors corresponding to features $i$ and $j$, and $k$ is the number of latent factors.

\begin{equation}
\mathbf{W}_{ij} = \langle \mathbf{v}_i, \mathbf{v}_j \rangle = \sum_{f=1}^{k} v_{if} v_{jf}
\label{eqn:interaction_weights}
\end{equation}

\noindent These interaction weights reveal the underlying dynamics of asset pairs, illustrating how they jointly influence the model’s predictions.

\noindent \textbf{Combined Attribution Score:} To enable comparison with other model-agnostic attribution methods, we combine linear and interaction effects into a single score as provided in Equation \eqref{eqn:combined_weights}, where $\alpha$ and $\beta$ are scaling factors that balance the contributions of linear and interaction weights.

\begin{equation}
\mathbf{w}_{\text{combined}} = \alpha \mathbf{w} + \beta \sum_{j=1}^{n} \mathbf{W}_{:, j}
\label{eqn:combined_weights}
\end{equation}

\noindent The combined score enables attribution of the detected anomaly to specific features, allowing for direct comparison with other methods. To thoroughly understand the influence of different factors in anomaly detection, both the linear and interaction weights are analyzed separately using the full FM Equation \eqref{eqn:fm_prediction}:
\begin{equation}
\hat{y}(\mathbf{x}) = w_0 + \sum_{i=1}^n w_i x_i + \sum_{i=1}^n \sum_{j=i+1}^n \langle \mathbf{v}_i, \mathbf{v}_j \rangle x_i x_j
\label{eqn:fm_prediction}
\end{equation}
This approach distinguishes between the contributions of individual features and their interactions, rather than merging them. By using the complete FM equation, which includes both linear terms ($w_i x_i$) and interaction terms ($\langle \mathbf{v}_i, \mathbf{v}_j \rangle x_i x_j$), the model provides a detailed view of how each feature (and feature pair) affects the prediction. 

\section{Evaluation and Results}
\label{sec:evaluation_results}

\subsection{Research Questions}

We investigate the following research questions on anomaly attribution and explainability in water systems: 

\begin{itemize}
    \item \textbf{RQ1:} How does FM improve explainability by modeling both individual and interaction effects in water systems?
    \item \textbf{RQ2:} What are the benefits of incorporating both linear and quadratic interactions in anomaly attribution for water system data?
    \item \textbf{RQ3:} How does FM compare to other model-agnostic attribution methods in terms of ranking and identifying the most influential features responsible for anomalies?
\end{itemize}

\begin{figure}[t]
    \centering
    \includegraphics[width=\linewidth]{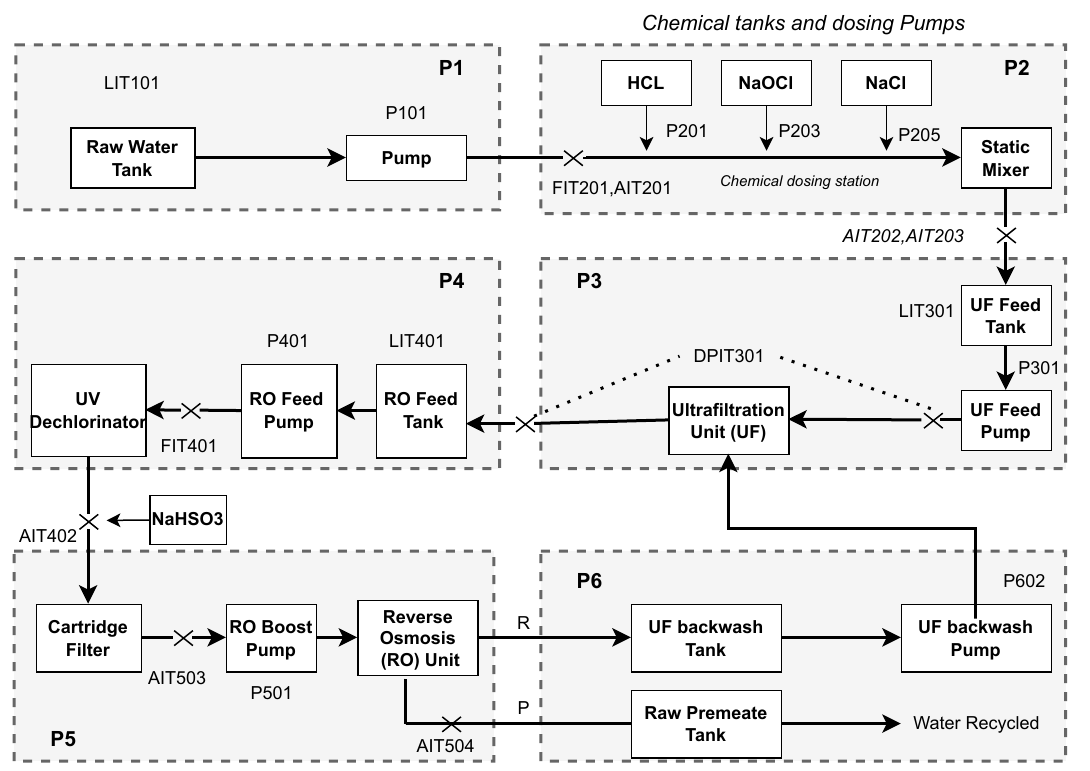}
    \caption{Overview of the \texttt{SWaT} System: The figure illustrates the architecture of the Secure Water Treatment (SWaT) system, highlighting the different stages along with key sensors and actuators involved in the treatment process. Here, \textit{R} stands for Reject and \textit{P} stands for Permeate.}
    \label{fig:swat_overview}
        \vspace{-0.3cm}
\end{figure}

\subsection{Datasets Used for Training and Evaluation}

We evaluate our method using two well-established datasets: \texttt{SWaT} \cite{goh2016secure} and \texttt{WADI} \cite{ahmed2017wadi}. Both are public datasets collected from real-world water systems in Singapore; frequently used in the anomaly detection research community \cite{mathuros2024}.

\noindent \textbf{SWaT Dataset.} This dataset, introduced by Goh et al. \cite{goh2016secure}, represents data collected from a scaled-down industrial \textit{water treatment plant}. This system simulates real-world setups with 6 distinct stages, in total equipped with 51 sensors (e.g., flow meters, pressure sensors) and 11 actuators (e.g., pumps, valves) as presented on Figure \ref{fig:swat_overview}. The dataset includes 7 days of normal operations and 4 days of attack scenarios, during which 41 carefully crafted cyber attacks, including 13 multi-feature and 28 single-feature attacks, were launched against the system. These attacks involved disruptions like spoofing sensor values and issuing malicious SCADA commands. Each device in the \texttt{SWaT} system is uniquely labeled to indicate its type and process stage; for example, `FIT-101' refers to a flow meter in the first process stage. This standardized labeling simplifies analysis and feature selection when applying machine learning models. The \texttt{SWaT} dataset has become a popular benchmark for testing anomaly detection algorithms and is widely used by researchers and practitioners\cite{Fung2024}.

\noindent \textbf{WADI Dataset.} This dataset is another public dataset collected from a \textit{water distribution system}, designed to replicate the operational environment of a real-world urban water network \cite{ahmed2017wadi}. \texttt{WADI} consists of a three-stage water distribution process controlled by Programmable Logic Controllers (PLCs) and Remote Terminal Units (RTUs) to manage sensors and actuators. The dataset includes two traces: one collected during normal operations and another containing multiple sequential attack scenarios. A total of 15 cyber-attacks were performed, including 6 multi-feature and 9 single-feature attacks such as sensor spoofing and actuator manipulation. The attacks were conducted by system operators, emulating various malicious activities like unauthorized actuator commands and false sensor readings.

\subsection{Evaluation Metrics}
\label{subsec:evaluation_metrics}

We employ the following metrics to evaluate the performance of our framework:

\noindent \textbf{\textit{AvgRank}}: Measures how effectively the model ranks the manipulated feature during an attack, with lower AvgRank indicating better performance. It is calculated by sorting features based on attribution scores and normalizing the manipulated feature's rank by the total number of features. For example, an AvgRank value of 0.2 means the manipulated feature is, on average, ranked in the top 20\% of all features. AvgRank is particularly useful for comparing our method with others (\texttt{LIME}, \texttt{SHAP} and \texttt{LEMNA}), as it quantifies the precision of feature rankings across methods.

\noindent \textbf{\textit{Top-k Feature Accuracy}:} Assesses whether the attacked feature is among the top k features identified by the attribution method. A high top-k accuracy indicates effective identification of the key contributing features.

\subsection{Detection Model Training and Setup}

Part of our approach relies on anomaly detection models. To this end, we train two distinct deep learning models: Convolutional Neural Networks (CNN) and Long Short-Term Memory (LSTM) networks. Although anomaly detection is not the primary focus of our study, we ensure robust detections by following the configurations outlined in \cite{Fung2024}. Both models consist of two layers with 64 units each and utilize a history window of 50 time steps. The Adam optimizer is employed, and the datasets are split into 80\% for training and 20\% for validation.

In our experiments, we evaluated the CNN and LSTM detectors on the \texttt{SWaT} and \texttt{WADI} datasets prior to attribution. For the \texttt{SWaT} dataset, which contains a total of 41 attacks, the CNN-based detector identified 33 attacks, and the LSTM-based detector identified 34 attacks. Specifically, out of 13 multi-feature attacks in \texttt{SWaT}, 8 were successfully detected by both models. For the \texttt{WADI} dataset, consisting of 24 total attacks, both CNN and LSTM detectors identified 15 attacks each. Among these, only 1 multi-feature attack was detected in \texttt{WADI}, which was subsequently omitted from attribution analysis due to insufficient sample size. Due to the similar performance of CNN and LSTM detectors, we continue our evaluation using only the LSTM model.

Since our study focuses solely on the attribution aspect, false positives were filtered out from the evaluation of attribution performance to ensure that only correctly identified attacks contribute to the attribution analysis.






\subsection{Implementation Details}

In our experiments, the FM explainer model is trained using the following hyperparameters: the dimension of latent vectors is set to \( k = 10 \), and the regularization parameters \( \lambda_w \) and \( \lambda_V \) are both set to 1.0. To prevent overfitting, we apply L1 regularization and select \( k \) through multiple tests, ensuring that \( 1 < k \ll \) the number of features to maintain model simplicity without compromising performance.

At the detection point, Gaussian noise with variance \( \sigma^2 = 0.05 \) is added to the input samples, as described in the methodology. To compute the combined attribution score, we apply a scaling factor to balance the contributions of the linear and interaction weights. The combined weights are calculated using Equation~\eqref{eqn:combined_weights_scaled}, where \( \alpha \) and \( \beta \) are scaling factors, \( \mathbf{w} \) are the linear weights, \( \mathbf{W}_{\text{interaction}} \) are the interaction weights, \( \overline{\mathbf{w}} \) and \( \overline{\mathbf{W}} \) are the mean values of the linear and interaction weights, respectively, and \( n \) is the number of features.

\begin{equation}
\label{eqn:combined_weights_scaled}
\mathbf{w}_{\text{combined}} = \alpha \mathbf{w} + \beta \left( \frac{\overline{\mathbf{w}}}{\overline{\mathbf{W}} \cdot n} \sum_{j=1}^{n} \mathbf{W}_{:,j} \right)
\end{equation}

The scaling factor \( \frac{\overline{\mathbf{w}}}{\overline{\mathbf{W}} \cdot n} \) normalizes the interaction weights relative to the linear weights, accounting for the difference in their magnitudes due to the number of interactions. During evaluation, we perform attribution for detected attacks using a realistic timing strategy. Specifically, at the point of detection, we compute the attribution at each time step over the following 50 time steps. The attribution scores are then aggregated by averaging across these 50 time steps, providing a robust and reliable estimate of feature attributions post-anomaly detection.

\subsection{Comparing Attribution Methods}
\label{subsec:comparing_attributions}

We compare our FM-based method with contemporary model-agnostic attribution methods such as \texttt{LIME}, \texttt{SHAP}, and \texttt{LEMNA}. While these methods primarily focus on individual feature contributions, they tend to overlook second-order interaction effects, which can be critical in water systems. We evaluate FM with combinations of $\alpha=0.5$, $\beta=0.5$ (FM(0.5)) and $\alpha=0.3$, $\beta=0.7$ (FM(0.7)), as defined in Equation~\ref{eqn:combined_weights_scaled}.

\begin{table}[t]
\centering
\caption{Attack Attribution Performance on \texttt{SWaT} Dataset.}
\label{tab:swat_results}
\resizebox{\columnwidth}{!}{%
\begin{tabular}{@{}lcccccc@{}}
\toprule
\multirow{2}{*}{\textbf{Method}} & \multicolumn{3}{c}{\textbf{All Attacks}} & \multicolumn{3}{c}{\textbf{Multi-Feature Attacks}} \\
\cmidrule(lr){2-4} \cmidrule(lr){5-7}
 & \textbf{AvgRank} & \textbf{Top-1} & \textbf{Top-5} & \textbf{AvgRank} & \textbf{Top-1} & \textbf{Top-5} \\
\midrule
\rowcolor[HTML]{EFEFEF}FM(0.5)   & 0.3411 & \textbf{6.82\%} & 18.42\% & \textbf{0.1539} & 25.00\% & 35.00\% \\
FM(0.7)   & 0.3378 & \textbf{6.73\% }& 18.47\% & 0.1549 & \textbf{25.00\%} & \textbf{35.00\%} \\
\rowcolor[HTML]{EFEFEF}\texttt{SHAP}    & \textbf{0.2908} & 6.27\% & \textbf{21.27\%} & 0.1892 & 10.00\% & 35.00\% \\
\texttt{LEMNA}     & 0.3024 & 6.22\% & 20.42\% & 0.2010 & 10.00\% & 30.00\% \\
\rowcolor[HTML]{EFEFEF}\texttt{LIME}      & 0.3592 & 0.00\% & 10.20\% & 0.1706 & 0.00\% & 25.00\% \\
\bottomrule
\end{tabular}%
}    \vspace{-0.3cm}
\end{table}

\noindent \textbf{Results on \texttt{SWaT} Dataset.} Table~\ref{tab:swat_results} presents the performance metrics of multiple model-agnostic attribution methods on the \texttt{SWaT} dataset, evaluated across both aggregated and multi-feature attack scenarios. Among the evaluated methods, \texttt{SHAP} achieved the lowest \textit{AvgRank} (0.2908) and the highest \textit{Top-5} accuracy (21.27\%) across all attack types. However, both FM(0.5) and FM(0.7) outperformed all other methods in multi-feature attack scenarios, where multiple sensors or actuators are manipulated. For these multi-feature attacks, FM(0.5) and FM(0.7) achieved lower \textit{AvgRank} values of 0.1539 and 0.1549, compared to \texttt{SHAP}'s 0.1892 and \texttt{LEMNA}'s 0.2010. Both FM methods also reached a \textit{Top-1} accuracy of 25.00\%, significantly higher than \texttt{SHAP} and \texttt{LEMNA} (both 10.00\%). In terms of \textit{Top-5} accuracy, FM(0.5) and FM(0.7) scored 35.00\%, outperforming \texttt{LEMNA} (30.00\%) and matching \texttt{SHAP} (35.00\%). Although \texttt{LEMNA} performed competitively with \texttt{SHAP} in some metrics, it struggled with multi-feature attacks, as reflected by its higher \textit{AvgRank} and lower \textit{Top-1} and \textit{Top-5} accuracies. This demonstrates the FM Explainer's effectiveness in capturing complex feature interactions, improving its ability to identify manipulated features in multi-feature attack scenarios. In contrast, \texttt{LIME} struggled in both aggregated and multi-feature attack contexts, particularly in identifying key manipulated features. It showed low \textit{Top-1} and \textit{Top-5} accuracies, likely due to its linear nature and focus on modeling individual feature effects, limiting its ability to capture important interactions. Overall, the FM-based method showed significant improvements in multi-feature attack scenarios while delivering comparable results across all attack types. Its ability to model pairwise (quadratic) interactions, alongside linear contributions, provides greater explainability. The method is also flexible, as \(\alpha\) and \(\beta\) can be tuned to emphasize either individual features or interactions. All methods, including FM, are model-agnostic and operate in linear time, making this a fair comparison for practical use in attributing anomalies to attack targets and protecting critical water infrastructure from cyber-attacks.

\begin{table}[t]
    \centering
    \caption{Attack Attribution Performance Metrics on \texttt{WADI} Dataset (Single-Feature Attacks).}
    \resizebox{\columnwidth}{!}{
    \renewcommand{\arraystretch}{0.8}
    \label{tab:wadi_results}
    \begin{tabular}{lcccc}
        \toprule
        \textbf{Method} & \textbf{AvgRank} & \textbf{Top-1} & \textbf{Top-5} & \textbf{Top-10} \\
        \midrule
        \rowcolor[HTML]{EFEFEF}FM (0.5) & 0.1225 & 52.62\% & 74.38\% & \textbf{75.00\%} \\
        FM (0.7) & 0.1223 & 56.13\% & \textbf{74.62\%} & \textbf{75.00\%} \\
       \rowcolor[HTML]{EFEFEF}\texttt{SHAP} & 0.1539 & \textbf{70.67\%} & 71.75\% & 75.00\% \\
        \texttt{LEMNA} & 0.1174 & 69.50\% & 72.00\% & 72.50\% \\
       \rowcolor[HTML]{EFEFEF}\texttt{LIME} & 0.3664 & 0.00\% & 0.80\% & 2.50\% \\

        \bottomrule
    \end{tabular}
}    \vspace{-0.3cm}
\end{table}

\noindent \textbf{Results on \texttt{WADI} Dataset.} We evaluated the attribution performance of our FM Explainer method against \texttt{SHAP}, \texttt{LIME}, and \texttt{LEMNA} on the \texttt{WADI} dataset, focusing on single-feature attacks detected by the anomaly detection model. Table~\ref{tab:wadi_results} presents the performance metrics using a practical timing strategy, which evaluates attribution over the 50 timesteps following the detection event. As shown in Table~\ref{tab:wadi_results}, our FM Explainer achieves strong performance in single-feature attacks, delivering competitive AvgRank values. The best-performing FM variant ($\alpha = 0.3$, $\beta = 0.7$) achieves an AvgRank of 0.1223, closely matching \texttt{LEMNA}'s best AvgRank of 0.1174. This demonstrates that, just as in multi-feature attacks, FM effectively ranks the attacked features higher in single-feature scenarios, outperforming \texttt{SHAP} (AvgRank 0.1539) and significantly better than \texttt{LIME} (AvgRank 0.3664). 
At Top-1 accuracy, SHAP outperforms FM. However, this advantage is
confined to the Top-1 metric, as FM remains highly competitive or superior across
Top-5 and Top-10 metrics.
 Both FM and \texttt{LEMNA} consistently outperform \texttt{SHAP} and significantly outperform \texttt{LIME}, which struggles across all metrics. These results show that, beyond the performance boost in multi-feature attacks, our FM Explainer also performs well in single-feature attack scenarios. Additionally, it provides more detailed explainability by capturing both individual and interaction effects, which other methods may overlook.

\begin{figure*}[htbp]
    \centering
    \begin{subfigure}[b]{0.48\textwidth}  
        \includegraphics[width=\textwidth]{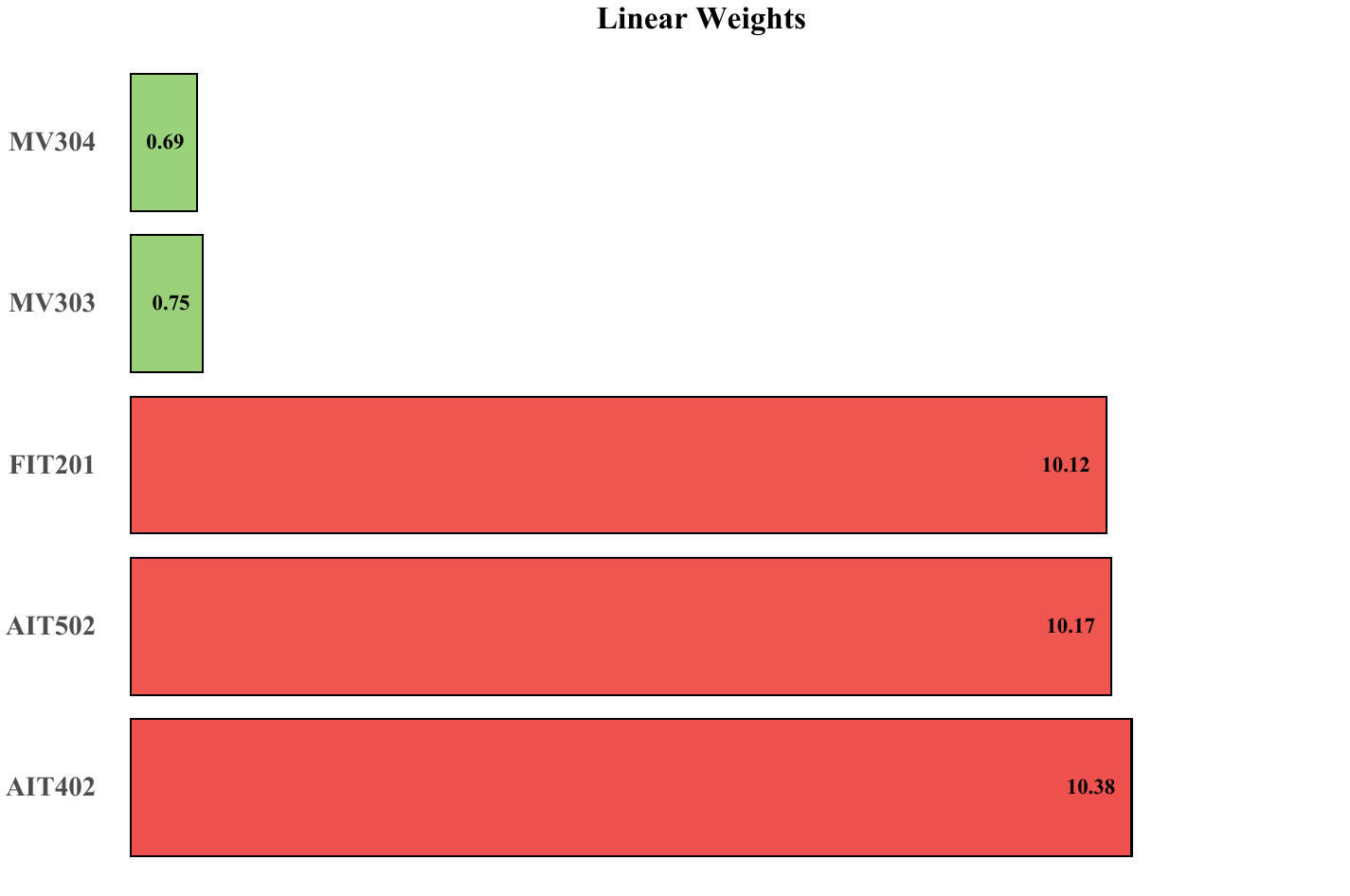}  
        \caption{Linear Weights}
        \label{fig:barplot}
    \end{subfigure}
    \hfill
    \begin{subfigure}[b]{0.48\textwidth}  
        \includegraphics[width=\textwidth]{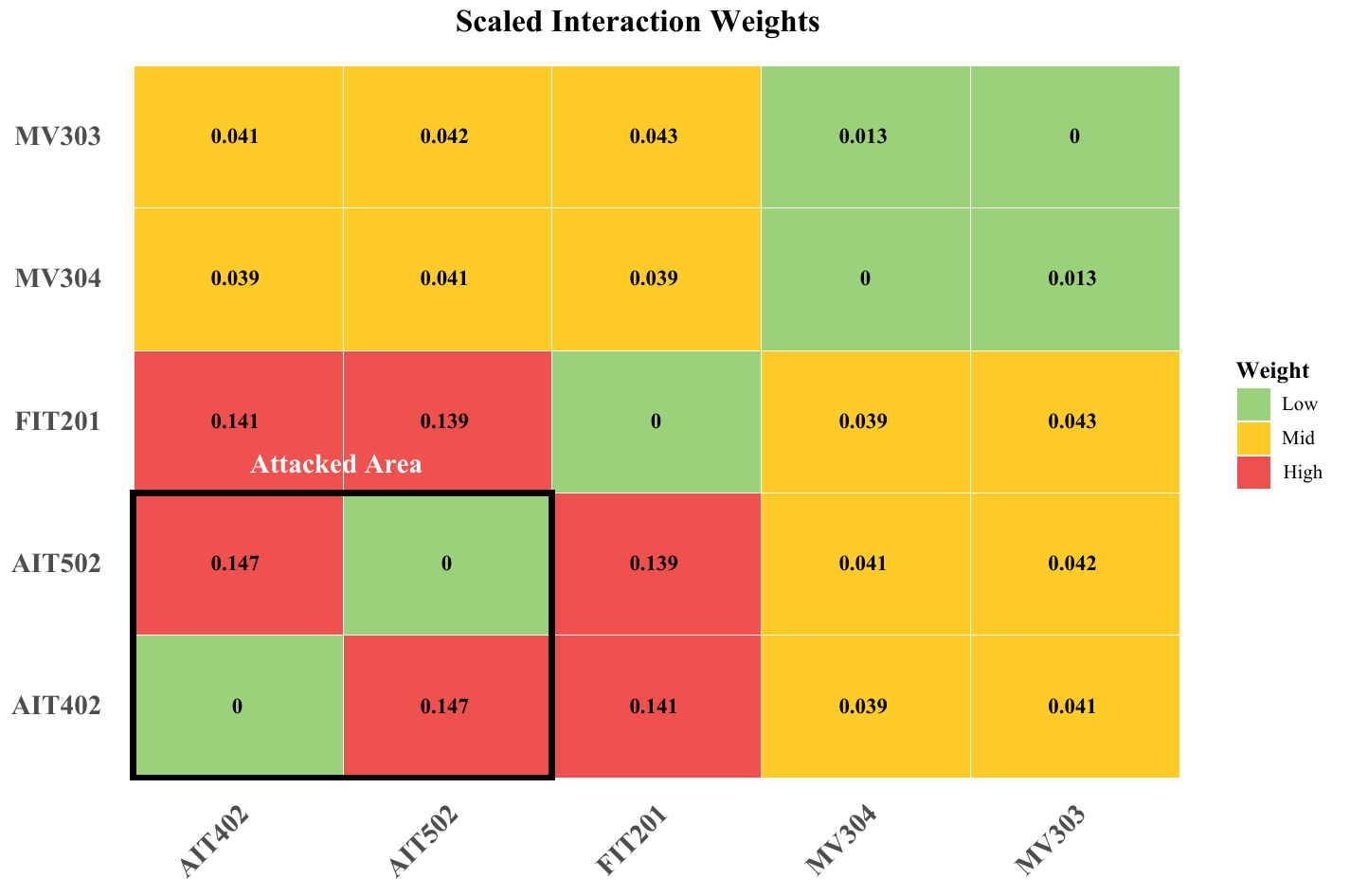}  
        \caption{Interaction Weights with Attacked Area}
        \label{fig:heatmap}
    \end{subfigure}
    
    \caption{(a) The Linear Weights bar plot on the left shows the individual contributions of key sensors and actuators to the anomaly. AIT402, AIT502, and FIT201 exhibit the highest weights, underscoring their critical roles in the detected anomaly. \\ (b) The Interaction Weights heatmap on the right illustrates the relationships between components, with the attacked area clearly highlighted through the interaction between AIT402 and AIT502. This provides further insight into how component interactions contribute to the anomaly.
}
    \label{fig:combined_weights}
    \vspace{-0.3cm}
\end{figure*}

 \subsection{Explaining Multi-Point Attacks in Water Systems}
\label{subsec:explainability_use_case}

This section demonstrates how our explainability framework identifies and analyzes multi-point attacks in the SWaT system by examining both individual feature contributions and their interactions. Figure~\ref{fig:combined_weights} shows the explainability results for Attack ID 27, where sensors \texttt{AIT402} and \texttt{AIT502} were manipulated. The left panel (a) shows the linear weights, representing individual feature contributions to the anomaly, while the right panel (b) displays interaction weights, highlighting dependencies between components. The linear weights reveal that \texttt{AIT402}, \texttt{AIT502}, and \texttt{FIT201} are the main contributors, consistent with the attack on \texttt{AIT402} and \texttt{AIT502}. A key advantage of our method over \texttt{LIME}, \texttt{SHAP}, and \texttt{LEMNA} is the ability to model interaction effects between features. The interaction heatmap highlights strong dependencies between \texttt{AIT402} and \texttt{AIT502}, confirming their joint role in the anomaly. Traditional methods, which focus on individual feature contributions, would likely miss these interactions since their outputs are not designed to capture feature dependencies. Considering both individual contributions and interactions enables more accurate identification of the components targeted by the attacker and how their relationships contribute to the anomaly. The adjustable parameters (\(\alpha\) and \(\beta\)) allow operators to to balance the emphasis between individual features and their interactions.

\section{Related Work}

Attribution methods seek to explain the predictions of machine learning models by estimating the contribution of each input feature to the output. For example, in image classification, attribution methods assign importance scores to pixels to interpret the model's decision~\cite{sundararajan2017axiomatic,Erion2021}. These methods are typically classified into two categories: model-agnostic and model-specific approaches.\\
\noindent \textbf{Model-agnostic attribution methods} rely solely on model inputs and outputs, without needing access to internal model parameters. Techniques like \texttt{LIME}~\cite{Ribeiro2016LIME} and \texttt{SHAP}~\cite{lundberg2017unified} generate local explanations by perturbing inputs and observing the resulting changes in outputs. \texttt{LIME} fits a simple linear model to approximate the complex model locally, using feature coefficients as attribution scores, while \texttt{SHAP} leverages Shapley values from cooperative game theory to distribute the contribution of each feature. More recent model-agnostic methods, such as \texttt{LEMNA}~\cite{guo2018lemna}, extend this approach by combining fused Lasso regression with Gaussian mixture models, making them more suitable for security-relevant applications. These methods have already been applied in ICS anomaly detection~\cite{Antwarg2021, Hwang2021}, offering valuable insights into which features contribute most to the detected anomalies. \\
\noindent \textbf{Model-specific attribution methods} rely on internal model gradients to understand how changes in input features influence the output. Prominent methods include saliency maps~\cite{Simonyan2014}, \texttt{Integrated Gradients} ~\cite{sundararajan2017axiomatic}, \texttt{SmoothGrad}~\cite{smilkov2017smoothgrad}, and \texttt{Expected Gradients}~\cite{Erion2021}, which have enhanced attribution quality, particularly in deep learning models, by leveraging gradients to highlight the most influential features. On the other hand, methods like \texttt{DeepLIFT}~\cite{shrikumar2017learning} introduced finer granularity in feature attribution, offering more detailed explanations. These approaches are foundational in the development of model-specific attribution techniques. In water systems, explainability and attribution for anomaly detection have been explored in prior work. Kravchik and Shabtai~\cite{Kravchik2022} used a CNN-based model on the \texttt{SWaT} dataset to identify anomalies through high-error features, while Hwang and Lee~\cite{Hwang2021} applied \texttt{SHAP} with an LSTM-based model. More advanced methods, such as \texttt{DeepSHAP}~\cite{lundberg2018explainable}, have shown promise in capturing complex relationships in cyber-physical systems. Model-agnostic attribution methods often overlook feature interactions, which are inherent in water systems due to the interconnected nature of sensors and actuators. This focus on individual features leads to incomplete explanations. To address this, we propose a model-agnostic solution that works across different anomaly detection models, recognizing that deep learning systems are frequently updated or replaced. Our approach uses FM~\cite{Rendle2010FM} for flexibility across different anomaly detection models while capturing both individual contributions and feature interactions. Compared to model-agnostic methods like \texttt{LIME}, \texttt{SHAP}, and \texttt{LEMNA}, FM offers greater accuracy in anomaly attribution, particularly in multi-component attacks. Its strength lies in modeling both linear and interaction effects, providing deeper insights into system behavior.

\section{Concluding remarks and future directions}
\label{sec:conclusion_future_work}

In this work, we introduce a novel model-agnostic Factorization Machines (FM)-based attack attribution method specifically tailored for water systems. This approach effectively addresses the limitations of existing attribution methods, including \texttt{LIME}, \texttt{SHAP}, and \texttt{LEMNA}, by simultaneously modeling both individual contributions and feature interactions. Our experiments on two real-world water system datasets, SWaT and WADI, show that the FM explainer outperforms existing methods in multi-feature attacks while delivering strong, comparable performance in single-feature attack scenarios. Notably, our approach is computationally efficient, enabling it to scale effectively to complex and interconnected water systems. This scalability is crucial for practical applications in real-time anomaly detection and attack attribution within critical infrastructure. In our future research endeavors, we aim to advance our work by expanding water system-related datasets to include more complex and stealthy attack scenarios. Additionally, we plan to enhance our FM-based attribution method by incorporating higher-order interactions, allowing us to estimate group or process stage contributions in water systems, and better assess the effect of group interconnectedness on detected anomalies caused by attacks. Recognizing the significant potential for enhancement in existing deep learning detection frameworks, our goal is to harness the capabilities of our FM approach to create a comprehensive solution that seamlessly integrates anomaly detection with attack attribution.

\section*{Acknowledgments}
\addcontentsline{toc}{section}{Acknowledgment}
The authors would like to thank the anonymous reviewer for their constructive comments. This work was supported by NSF under Grants \#2230087 and \#2404946.

\bibliographystyle{IEEEtran} 
\bibliography{ref} 

\begin{thebibliography}{10}
\providecommand{\url}[1]{#1}
\csname url@samestyle\endcsname
\providecommand{\newblock}{\relax}
\providecommand{\bibinfo}[2]{#2}
\providecommand{\BIBentrySTDinterwordspacing}{\spaceskip=0pt\relax}
\providecommand{\BIBentryALTinterwordstretchfactor}{4}
\providecommand{\BIBentryALTinterwordspacing}{\spaceskip=\fontdimen2\font plus
\BIBentryALTinterwordstretchfactor\fontdimen3\font minus
  \fontdimen4\font\relax}
\providecommand{\BIBforeignlanguage}[2]{{%
\expandafter\ifx\csname l@#1\endcsname\relax
\typeout{** WARNING: IEEEtran.bst: No hyphenation pattern has been}%
\typeout{** loaded for the language `#1'. Using the pattern for}%
\typeout{** the default language instead.}%
\else
\language=\csname l@#1\endcsname
\fi
#2}}
\providecommand{\BIBdecl}{\relax}
\BIBdecl

\bibitem{umich2023water}
{Center for Sustainable Systems, University of Michigan}, ``U.s. water supply
  and distribution factsheet,''
  \url{http://css.umich.edu/factsheets/us-water-supply-and-distribution-factsheet},
  2023, pub. No. CSS05-17.

\bibitem{smartcity}
N.~Neshenko, C.~Nader, E.~Bou-Harb, and B.~Furht, ``A survey of methods
  supporting cyber situational awareness in the context of smart cities,''
  \emph{Journal of Big Data}, vol.~7, 10 2020.

\bibitem{kans}
Bleepingcomputer, ``Kansas water plant cyberattack forces switch to manual
  operations,''
  \url{https://www.bleepingcomputer.com/news/security/kansas-water-plant-cyberattack-forces-switch-to-manual-operations/},
  2024, accessed: 2024-09-30.

\bibitem{goodin2021florida}
D.~Goodin, ``Hackers tried to poison water supply of florida town by remotely
  accessing treatment plant controls,''
  \url{https://arstechnica.com/information-technology/2021/02/hackers-try-to-poison-water-supply-of-florida-town/},
  2021, accessed: 2023-10-15.

\bibitem{tucker2020israel}
P.~Tucker, ``Israeli cyber officials: Water-systems attack was coordinated iran
  effort,''
  \url{https://www.defenseone.com/threats/2020/05/israeli-cyber-officials-water-systems-attack-was-coordinated-iran-effort/165387/},
  2020, accessed: 2023-10-15.

\bibitem{kim2023iran}
\BIBentryALTinterwordspacing
J.~Kim, ``Iran-linked cyberattacks threaten equipment used in u.s. water
  systems and factories,'' \emph{NPR}, 2023, accessed: 2024-10-07. [Online].
  Available:
  \url{https://www.npr.org/2023/12/02/1216735250/iran-linked-cyberattacks-israeli-equipment-water-plants}
\BIBentrySTDinterwordspacing

\bibitem{eliascps}
E.~Bou-Harb, ``A brief survey of security approaches for cyber-physical
  systems,'' in \emph{2016 8th IFIP International Conference on New
  Technologies, Mobility and Security (NTMS)}, 2016, pp. 1--5.

\bibitem{Fung2024}
\BIBentryALTinterwordspacing
C.~Fung, E.~Zeng, and L.~Bauer, ``Attributions for ml-based ics anomaly
  detection: From theory to practice,'' in \emph{Network and Distributed System
  Security (NDSS) Symposium 2024}.\hskip 1em plus 0.5em minus 0.4em\relax San
  Diego, CA, USA: Internet Society, February 26 - March 1 2024. [Online].
  Available: \url{https://dx.doi.org/10.14722/ndss.2024.23216}
\BIBentrySTDinterwordspacing

\bibitem{moraitis2022cyberphysical}
G.~Moraitis, I.~Tsoukalas, P.~Kossieris, D.~Nikolopoulos, G.~Karavokiros,
  D.~Kalogeras, and C.~Makropoulos, ``Assessing cyber-physical threats under
  water demand uncertainty,'' \emph{Environ. Sci. Proc.}, vol.~21, no.~1,
  p.~18, 2022.

\bibitem{taormina2017characterizing}
R.~Taormina, S.~Galelli, N.~O. Tippenhauer, E.~Salomons, and A.~Ostfeld,
  ``Characterizing cyber-physical attacks on water distribution systems,''
  \emph{Journal of Water Resources Planning and Management}, vol. 143, no.~5,
  p. 04017009, 2017.

\bibitem{giraldo2018survey}
J.~Giraldo, D.~Urbina, A.~Cardenas, J.~Valente, M.~Faisal, J.~Ruths, N.~O.
  Tippenhauer, H.~Sandberg, and R.~Candell, ``A survey of physics-based attack
  detection in cyber-physical systems,'' \emph{ACM Computing Surveys (CSUR)},
  vol.~51, no.~4, pp. 1--36, 2018.

\bibitem{ha2022explainable}
D.~T. Ha, N.~X. Hoang, N.~V. Hoang, N.~H. Du, T.~T. Huong, and K.~P. Tran,
  ``Explainable anomaly detection for industrial control system
  cybersecurity,'' \emph{IFAC-PapersOnLine}, vol.~55, no.~10, pp. 103--108,
  2022.

\bibitem{mitreT1565}
\BIBentryALTinterwordspacing
MITRE, ``Data manipulation - mitre att\&ck technique t1565,'' 2023, accessed:
  2024-10-07. [Online]. Available:
  \url{https://attack.mitre.org/techniques/T1565/}
\BIBentrySTDinterwordspacing

\bibitem{ahmed2017wadi}
C.~S. Ahmed, V.~A. Palleti, and A.~Mathur, ``Wadi: A water distribution testbed
  for research in the design of secure cyber physical systems,''
  \emph{Proceedings of the 3rd international workshop on cyber-physical systems
  for smart water networks}, pp. 25--28, 2017.

\bibitem{khoury2020hybrid}
J.~Khoury and M.~Nassar, ``A hybrid game theory and reinforcement learning
  approach for cyber-physical systems security,'' in \emph{NOMS 2020-2020
  IEEE/IFIP Network Operations and Management Symposium}.\hskip 1em plus 0.5em
  minus 0.4em\relax IEEE, 2020, pp. 1--9.

\bibitem{alsabaan2023}
M.~Alsabaan, M.~I. Ibrahem, and H.~Elwahsh, ``Real-time anomaly detection for
  water quality sensor monitoring based on multivariate deep learning
  technique,'' \emph{Sensors}, vol.~23, no.~20, 2023.

\bibitem{yang2022}
C.~Yang, R.~Paepae, and F.~Zhang, ``A survey on applications of machine
  learning algorithms in water quality assessment and water supply
  management,'' \emph{Water Supply}, 2022.

\bibitem{ahmed2023}
K.~Ahmed and S.~Singh, ``Application of machine learning in river water quality
  management: a review,'' \emph{Water Science and Technology}, 2023.

\bibitem{Joseph}
J.~Khoury, {\DJ}.~Klisura, H.~Zanddizari, G.~De~La Torre~Parra, P.~Najafirad,
  and E.~Bou-Harb, ``Jbeil: Temporal graph-based inductive learning to infer
  lateral movement in evolving enterprise networks,'' in \emph{2024 IEEE
  Symposium on Security and Privacy (SP)}, 2024, pp. 3644--3660.

\bibitem{goh2016dataset}
J.~Goh, S.~Adepu, M.~Tan, Z.~K. Lee, and A.~P. Mathur, ``A dataset to support
  research in the design of secure water treatment systems,'' in
  \emph{International Conference on Critical Information Infrastructures
  Security}.\hskip 1em plus 0.5em minus 0.4em\relax Springer, 2016, pp. 88--99.

\bibitem{fung2022perspectives}
C.~Fung, S.~Srinarasi, K.~Lucas, H.~B. Phee, and L.~Bauer, ``Perspectives from
  a comprehensive evaluation of reconstruction-based anomaly detection in
  industrial control systems,'' in \emph{27th European Symposium on Research in
  Computer Security}.\hskip 1em plus 0.5em minus 0.4em\relax Springer, 2022.

\bibitem{zizzo2019intrusion}
G.~Zizzo, C.~Hankin, S.~Maffeis, and K.~Jones, ``Intrusion detection for
  industrial control systems: Evaluation analysis and adversarial attacks,''
  \emph{arXiv preprint arXiv:1911.04278}, 2019.

\bibitem{hwang2021esfd}
C.~Hwang and T.~Lee, ``E-sfd: Explainable sensor fault detection in the ics
  anomaly detection system,'' \emph{IEEE Access}, vol.~9, pp.
  140\,470--140\,486, 2021.

\bibitem{Ribeiro2016LIME}
\BIBentryALTinterwordspacing
M.~T. Ribeiro, S.~Singh, and C.~Guestrin, ``“why should i trust you?”
  explaining the predictions of any classifier,'' in \emph{Proceedings of the
  22nd ACM SIGKDD International Conference on Knowledge Discovery and Data
  Mining (KDD '16)}.\hskip 1em plus 0.5em minus 0.4em\relax New York, NY, USA:
  ACM, 2016, pp. 1135--1144. [Online]. Available:
  \url{https://doi.org/10.1145/2939672.2939778}
\BIBentrySTDinterwordspacing

\bibitem{lundberg2017unified}
S.~M. Lundberg and S.-I. Lee, ``A unified approach to interpreting model
  predictions,'' in \emph{Advances in Neural Information Processing Systems},
  2017, pp. 4765--4774.

\bibitem{guo2018lemna}
W.~Guo, D.~Mu, J.~Xu, P.~Su, G.~Wang, and X.~Xing, ``Lemna: Explaining deep
  learning based security applications,'' in \emph{Proceedings of the 2018 ACM
  SIGSAC Conference on Computer and Communications Security}, 2018, pp.
  364--379.

\bibitem{sundararajan2017axiomatic}
M.~Sundararajan, A.~Taly, and Q.~Yan, ``Axiomatic attribution for deep
  networks,'' in \emph{34th International Conference on Machine Learning},
  2017, pp. 3319--3328.

\bibitem{smilkov2017smoothgrad}
D.~Smilkov, N.~Thorat, B.~Kim, F.~Viegas, and M.~Wattenberg, ``Smoothgrad:
  Removing noise by adding noise,'' \emph{arXiv preprint arXiv:1706.03825},
  2017.

\bibitem{simonyan2013deep}
K.~Simonyan, A.~Vedaldi, and A.~Zisserman, ``Deep inside convolutional
  networks: Visualising image classification models and saliency maps,''
  \emph{arXiv preprint arXiv:1312.6034}, 2013.

\bibitem{sadegh}
S.~Torabi, M.~Dib, E.~Bou-Harb, C.~Assi, and M.~Debbabi, ``A strings-based
  similarity analysis approach for characterizing iot malware and inferring
  their underlying relationships,'' \emph{IEEE Networking Letters}, vol.~3,
  no.~3, pp. 161--165, 2021.

\bibitem{lecun1998gradient}
Y.~LeCun, L.~Bottou, Y.~Bengio, and P.~Haffner, ``Gradient-based learning
  applied to document recognition,'' \emph{Proceedings of the IEEE}, vol.~86,
  no.~11, pp. 2278--2324, 1998.

\bibitem{hochreiter1997long}
S.~Hochreiter and J.~Schmidhuber, ``Long short-term memory,'' \emph{Neural
  computation}, vol.~9, no.~8, pp. 1735--1780, 1997.

\bibitem{Rendle2010FM}
\BIBentryALTinterwordspacing
S.~Rendle, ``Factorization machines,'' in \emph{Proceedings of the 10th IEEE
  International Conference on Data Mining (ICDM '10)}.\hskip 1em plus 0.5em
  minus 0.4em\relax Washington, DC, USA: IEEE Computer Society, 2010, pp.
  995--1000. [Online]. Available: \url{https://doi.org/10.1109/ICDM.2010.127}
\BIBentrySTDinterwordspacing

\bibitem{Rendle2012FMlibFM}
------, ``Factorization machines with libfm,'' \emph{ACM Transactions on
  Intelligent Systems and Technology (TIST)}, vol.~3, no.~3, pp. 1--22, 2012.

\bibitem{LI2021204}
\BIBentryALTinterwordspacing
J.~Li, C.~Sun, and Q.~Su, ``Analysis of cascading failures of power
  cyber-physical systems considering false data injection attacks,''
  \emph{Global Energy Interconnection}, vol.~4, no.~2, pp. 204--213, 2021.
  [Online]. Available:
  \url{https://www.sciencedirect.com/science/article/pii/S2096511721000402}
\BIBentrySTDinterwordspacing

\bibitem{antonio_elias}
\BIBentryALTinterwordspacing
A.~Mangino, M.~S. Pour, and E.~Bou-Harb, ``Internet-scale insecurity of
  consumer internet of things: An empirical measurements perspective,''
  \emph{ACM Trans. Manage. Inf. Syst.}, vol.~11, no.~4, Oct. 2020. [Online].
  Available: \url{https://doi.org/10.1145/3394504}
\BIBentrySTDinterwordspacing

\bibitem{apruzzese2023machine}
G.~Apruzzese, P.~Laskov, E.~Montes~de Oca, W.~Mallouli, L.~Brdalo~Rapa, A.~V.
  Grammatopoulos, and F.~Di~Franco, ``The role of machine learning in
  cybersecurity,'' \emph{Digital Threats: Research and Practice}, vol.~4,
  no.~1, 2023.

\bibitem{kravchik2022efficient}
M.~Kravchik and A.~Shabtai, ``Efficient cyber attack detection in industrial
  control systems using lightweight neural networks and pca,'' \emph{IEEE
  Transactions on Dependable and Secure Computing}, vol.~19, no.~4, pp.
  2455--2468, 2022.

\bibitem{goh2016secure}
J.~Goh, S.~Adepu, S.~Nadas \emph{et~al.}, ``Secure water treatment (swat): A
  testbed for research and training on ics security,'' \emph{IFIP International
  Information Security Conference}, pp. 235--239, 2016.

\bibitem{mathuros2024}
K.~Mathuros, S.~Venugopalan, and S.~Adepu, ``Waxai: Explainable anomaly
  detection in industrial control systems and water systems,'' \emph{CPSS'24},
  2024.

\bibitem{Erion2021}
G.~Erion, J.~D. Janizek, P.~Sturmfels, S.~M. Lundberg, and S.-I. Lee,
  ``Improving performance of deep learning models with axiomatic attribution
  priors and expected gradients,'' \emph{Nature Machine Intelligence}, vol.~3,
  no.~7, pp. 620--631, 2021.

\bibitem{Antwarg2021}
L.~Antwarg, R.~M. Miller, B.~Shapira, and L.~Rokach, ``Explaining anomalies
  detected by autoencoders using shapley additive explanations,'' \emph{Expert
  Systems with Applications}, vol. 186, p. 115736, 2021.

\bibitem{Hwang2021}
C.~Hwang and T.~Lee, ``E-sfd: Explainable sensor fault detection in the ics
  anomaly detection system,'' \emph{IEEE Access}, vol.~9, pp.
  140\,470--140\,486, 2021.

\bibitem{Simonyan2014}
K.~Simonyan, A.~Vedaldi, and A.~Zisserman, ``Deep inside convolutional
  networks: Visualising image classification models and saliency maps,''
  \emph{arXiv preprint arXiv:1312.6034}, 2013.

\bibitem{shrikumar2017learning}
A.~Shrikumar, P.~Greenside, and A.~Kundaje, ``Learning important features
  through propagating activation differences,'' in \emph{Proceedings of the
  34th International Conference on Machine Learning-Volume 70}.\hskip 1em plus
  0.5em minus 0.4em\relax JMLR.org, 2017, pp. 3145--3153.

\bibitem{Kravchik2022}
M.~Kravchik and A.~Shabtai, ``Efficient cyber attack detection in industrial
  control systems using lightweight neural networks and pca,'' \emph{IEEE
  Transactions on Dependable and Secure Computing}, vol.~19, no.~4, pp.
  2148--2164, 2022.

\bibitem{lundberg2018explainable}
S.~M. Lundberg and S.-I. Lee, ``Explainable ai for trees: From local
  explanations to global understanding,'' in \emph{arXiv preprint
  arXiv:1905.04610}, 2018.

\end{thebibliography}

\end{document}